%% file: main_ieee.tex
\documentclass[conference]{IEEEtran}

\usepackage{cite}
\usepackage{amsmath,amssymb,amsfonts}
\usepackage{textcomp}

\input{section/preamble}

\begin{document}

\title{Multitask Learning for Time Series Data\\with 2D Convolution}

\input{section/author}

\maketitle

\input{section/abstract}

\input{section/introduction}
\input{section/related}
\input{section/method}
\input{section/experiment}
\input{section/conclusion}

\Urlmuskip=0mu plus 1mu\relax
\bibliographystyle{IEEEtran}
\bibliography{section/ref}

\end{document}

%% file: section/preamble.tex
\usepackage{cite}
\usepackage{amsmath}
\usepackage{amssymb}
\usepackage{amsfonts}
\usepackage{hyperref}

\usepackage{algorithm}
\usepackage[noend]{algpseudocode}

\newcommand*\Input[1]{\Statex \textbf{Input:} #1}
\newcommand*\Output[1]{\Statex \textbf{Output:} #1}
\algrenewcommand\alglinenumber[1]{#1}
\algrenewcommand\algorithmicindent{1em}

\usepackage{amsthm}
\usepackage{caption}
\usepackage{subcaption}

\usepackage{booktabs}

\newtheoremstyle{def_style}
  {}          
  {}          
  {}          
  {}          
  {\bfseries} 
  {.}         
  {.5em}      
  {}          

\theoremstyle{def_style}

\theoremstyle{def_style}

\usepackage{multirow}

\usepackage{mathtools}

%% file: section/author.tex




\author{\IEEEauthorblockN{Chin-Chia Michael Yeh, Xin Dai, Yan Zheng, Junpeng Wang, Huiyuan Chen,\\Yujie Fan, Audrey Der$^\dagger$, Zhongfang Zhuang, Liang Wang, and Wei Zhang}
\IEEEauthorblockA{\textit{Visa Research}, \textit{University of California, Riverside}$^\dagger$ \\
\{miyeh,xidai,yazheng,junpenwa,hchen,yufan,audder,zzhuang,liawang,wzhan\}@visa.com}
}


%% file: section/abstract.tex
\begin{abstract}
Multitask learning (MTL) aims to develop a unified model that can handle a set of closely related tasks simultaneously. 
By optimizing the model across multiple tasks, MTL generally surpasses its non-MTL counterparts in terms of generalizability.
Although MTL has been extensively researched in various domains such as computer vision, natural language processing, and recommendation systems, its application to time series data has received limited attention. 
In this paper, we investigate the application of MTL to the time series classification (TSC) problem.
However, when we integrate the state-of-the-art $1D$ convolution-based TSC model with MTL, the performance of the TSC model actually deteriorates. 
By comparing the $1D$ convolution-based models with the Dynamic Time Warping (DTW) distance function, it appears that the underwhelming results stem from the limited expressive power of the $1D$ convolutional layers.
To overcome this challenge, we propose a novel design for a $2D$ convolution-based model that enhances the model's expressiveness. 
Leveraging this advantage, our proposed method outperforms competing approaches on both the UCR Archive and an industrial transaction TSC dataset.
\end{abstract}

\begin{IEEEkeywords}
time series, multitask learning, convolutional neural network
\end{IEEEkeywords}


%% file: section/introduction.tex
\section{Introduction}
Multitask learning (MTL) is a learning strategy that enhances a model's generalizability by leveraging shared knowledge from closely related tasks~\cite{ruder2017overview,liu2019multi,dai2021fishnet}. 
This approach has demonstrated significant success in domains such as computer vision~\cite{dai2021fishnet} and natural language processing~\cite{liu2019multi}.
Time series data are prevalent in various domains~\cite{dau2019ucr}, yet there is a surprising lack of MTL research focused on this type of data, despite its potential for transferability to time series analysis.
Consequently, this paper aims to investigate the application of MTL to time series classification (TSC).
Our study specifically concentrates on the widely adopted MTL technique known as hard parameter sharing. 
In this approach, different tasks share a common sub-network as a feature extractor, while each task maintains its own task-specific output layer. 
Hard parameter sharing is considered one of the most efficient and robust methods for implementing MTL, frequently serving as a baseline in many studies~\cite{misra2016cross,gao2019nddr,gao2020mtl,liu2019end,sun2019adashare}.

We can easily combine hard parameter sharing with one of the strong deep learning TSC baselines, such as \texttt{ResNet}~\cite{wang2017time,ismail2019deep}. 
However, when we combine hard parameter sharing with \texttt{ResNet}, we observe that 15 out of 25 tasks in our time series MTL benchmark datasets perform worse than their non-MTL counterparts (see Section~\ref{sec:exp:ucr} for details). 
Furthermore, we compare it with a simpler baseline: the 1-nearest neighbor algorithm with dynamic time warping distance\footnote{The \texttt{1NN+DTW} method is an effective approach for multiple TSC datasets/tasks~\cite{bagnall2017great}.} (\texttt{1NN+DTW})~\cite{rakthanmanon2012searching}. 
Surprisingly, we find that even \texttt{1NN+DTW} can outperform \texttt{ResNet} with MTL in terms of overall performance in our benchmark. 
This suggests that the warping mechanism (DTW) can easily capture some simple yet potentially useful features for MTL, which are challenging for the $1D$ convolutional layers of the \texttt{ResNet} baseline model to learn.

To comprehend why $1D$ convolution encounters difficulties in learning the warping mechanism, it is necessary to explain the computation process of the \texttt{DTW} distance.
Given two time series, the first step involves computing the pair-wise distance matrix between all pairs of elements in the two input time series.
Next, a recursive function is employed to process the pair-wise distance matrix, yielding the distance value.
The inductive bias of $1D$ locality from the $1D$ convolution operation poses difficulties in effectively capturing both the pair-wise distance computation and the recursive function.
To address this limitation, we propose a novel $2D$ convolutional TSC model that is specifically designed to better capture the warping mechanism compared to the baseline \texttt{ResNet}~\cite{wang2017time,ismail2019deep}. 
This is achieved by 1) explicitly processing the pair-wise distance matrix and 2) employing $2D$ convolutional layers to learn diverse recursive functions.
Please note that since the baseline ResNet utilizes $1D$ convolutional layers, we will refer to it as \texttt{ResNet1D} to indicate the baseline method. 
Similarly, we will use \texttt{ResNet2D} to denote the proposed method, which incorporates $2D$ convolutional layers.

\input{insert/fig_toy}

To provide empirical evidence supporting the need for a new model to learn the warping mechanism, we conducted a comparative experiment to assess the capability of different models in approximating the \texttt{DTW} distance function.
Both the baseline \texttt{ResNet1D} model and the proposed \texttt{ResNet2D} model were trained to approximate the \texttt{DTW} distance function under various warping constraints (refer to Section~\ref{sec:exp:dtw} for detailed information).
We measured the error between the predicted distance from the models and the actual distance output by the distance function. 
As depicted in Figure~\ref{fig:toy}, the proposed \texttt{ResNet2D} model outperformed the baseline \texttt{ResNet1D} model in approximating the \texttt{DTW} functions.
Based on this observation, we extended the proposed \texttt{ResNet2D} model from the distance learning task to the MTL TSC task. 
When combining the proposed \texttt{ResNet2D} model with hard parameter sharing MTL, it achieved the highest accuracy across multiple tasks compared to alternative approaches.

\bigbreak
\noindent This paper makes the following contributions::
\begin{itemize}
    \item We demonstrate that combining hard parameter sharing with a conventional deep learning model (\texttt{ResNet1D}) is ineffective for the TSC problem.
    \item We propose a novel $2D$ convolution-based neural network (\texttt{ResNet2D}) that can learn the warping mechanism and achieves superior performance in MTL with hard parameter sharing compared to the baseline methods.
    \item We validate the effectiveness of the proposed \texttt{ResNet2D} model on both publicly available MTL TSC datasets and an industrial MTL TSC dataset.
\end{itemize}

%% file: insert/fig_toy.tex
\begin{figure}[ht]
\centerline{
\includegraphics[width=0.9\linewidth]{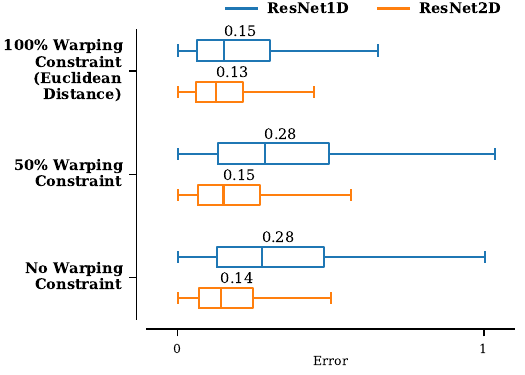}
}
\caption{
The proposed \texttt{ResNet2D} model can approximate DTW distance more accurately under various warping constraints compared to the baseline \texttt{ResNet1D} model.
}
\label{fig:toy}
\end{figure}

%% file: section/related.tex
\section{Related Work}
The two primary research areas relevant to our paper are \textit{multitask learning} (MTL) and \textit{time series classification} (TSC). 
In this section, we provide a brief literature review on both of research areas.

\subsection{Multitask Learning}
MTL is a learning paradigm that aims to enhance the generalizability of machine learning models~\cite{zhang2021survey,ruder2017overview}. 
It has gained significant attention in various domains, including computer vision, natural language processing, and recommendation systems~\cite{zhang2021survey,ruder2017overview}.
In recent years, efforts have been made to apply MTL to time series data~\cite{cirstea2018correlated,chen2021deep,chandra2017co,mcdermott2020comprehensive}. 
For instance, Cirstea et al.~\cite{cirstea2018correlated} proposed a hybrid convolutional-recurrent neural network for correlated time series forecasting, which incorporated a reconstruction task as a regularization term to improve generalization. 
Chen et al.~\cite{chen2021deep} introduced a representation learning method that jointly learns time series classification and retrieval.
However, these approaches assume that all tasks share a single dataset~\cite{cirstea2018correlated,chen2021deep,chandra2017co,mcdermott2020comprehensive}. In other words, these methods can only be applied to homogeneous-feature MTL problems~\cite{zhang2021survey}.
To the best of our knowledge, our proposed framework represents the first work addressing \textit{general} MTL on time series data.

The primary focus of MTL research lies in addressing the conflicts that arise among the objectives of different tasks~\cite{zhang2021survey}. 
Various studies have attempted to tackle this challenge by mitigating the conflicts in gradients associated with different losses~\cite{chen2018gradnorm,sener2018multi}.
For instance, Chen et al.~\cite{chen2018gradnorm} proposed a method to balance the gradients of multiple losses in a multi-task neural network. 
Sener et al.~\cite{sener2018multi} treated MTL as a multi-objective optimization problem and aimed to find a Pareto optimal solution.
Other research endeavors have focused on discovering effective parameter-sharing structures for multiple tasks~\cite{lu2017fully,guo2020learning,sun2019adashare}. 
For example, Lu et al.~\cite{lu2017fully} introduced a bottom-up searching algorithm to identify a suitable sharing structure for a given dataset. 
Guo et al.~\cite{guo2020learning} and Sun et al.~\cite{sun2019adashare} proposed approaches to learn the sharing structure directly from the dataset.

Soft parameter sharing MTL, which involves sharing feature maps instead of parameters, is another approach explored in computer vision research~\cite{misra2016cross,gao2019nddr,gao2020mtl}. 
For instance, Misra et al.~\cite{misra2016cross} proposed the \texttt{Cross-stitch} network, while Gao et al.~\cite{gao2019nddr} introduced the \texttt{NDDR-CNN}. 
These methods maintain task-specific networks but integrate their feature maps at intermediate layers~\cite{misra2016cross,gao2019nddr}.
Furthermore, Gao et al.~\cite{gao2020mtl} proposed \texttt{MTL-NAS}, a method that utilizes network architecture search techniques to explore potential sharing links between feature maps at various intermediate layers. 
Although our paper focuses on the basic approach of hard parameter sharing MTL, which has demonstrated robustness in preventing overfitting and serves as a common baseline in state-of-the-art studies~\cite{misra2016cross,gao2019nddr,gao2020mtl,liu2019end,sun2019adashare}, the ideas presented in the aforementioned studies offer interesting avenues for future research.

\subsection{Time Series Classification}
Over the past decades, numerous algorithms have been proposed for TSC, encompassing a wide range of approaches~\cite{bagnall2017great,ismail2019deep}. 
The spectrum includes simple yet effective methods such as the nearest neighbor classifier utilizing the \texttt{DTW} distance~\cite{dau2019ucr,bagnall2017great}, as well as complex and powerful ensemble techniques like \texttt{HIVE-COTE 2.0}~\cite{lines2015time,bagnall2017great,middlehurst2021hive}.
Inspired by the achievements of deep learning models in computer vision and natural language processing domains~\cite{he2016deep,devlin2018bert}, there have been numerous endeavors to adapt deep learning-based approaches for TSC problems~\cite{ismail2019deep}. 
For instance, Cui et al.~\cite{cui2016multi} proposed the multi-scale convolutional neural network (\texttt{MCNN}) model, which classifies input time series by employing multiple temporal scaling and smoothing operations with a convolutional neural network. 
Over the years, various variants of convolutional neural networks, including the \texttt{Time Le-Net} model\cite{le2016data}, \texttt{Time-CNN} model~\cite{zhao2017convolutional}, \texttt{FCN} model~\cite{wang2017time}, and \texttt{ResNet} model~\cite{wang2017time}, have been proposed for TSC.
Among these deep learning-based TSC models, \texttt{ResNet} continues to serve as a strong baseline, as demonstrated by extensive evaluations conducted by Ismail et al.~\cite{ismail2019deep}. 
Consequently, our paper aims to enhance the performance of the \texttt{ResNet} model specifically for the TSC problem within the MTL framework.

%% file: section/method.tex
\section{Methodology}
In this section, we present the design of two models: the baseline \texttt{ResNet1D} model and the proposed \texttt{ResNet2D} model. 
These models are used for both distance function approximation and classification tasks, and we provide an explanation of how each model is employed in each type of problem.
Furthermore, we discuss the optimization process employed for training these models.

\subsection{\texttt{ResNet1D} Architecture}
Figure~\ref{fig:resnet_1d} illustrates the design of the baseline \texttt{ResNet1D} model and how it is used in our experiments. 
The model design is inspired by the \texttt{ResNet} architecture proposed in~\cite{wang2017time}.

\input{insert/fig_resnet_1d}

The building block of the \texttt{ResNet1D} model is depicted in Figure~\ref{fig:resnet_1d}a.
Each $1D$ convolutional layer (\texttt{conv}) is annotated with its corresponding filter size.
The stride size for all \texttt{conv} layers is set to 1.
In cases where the input and output dimensions of the building block differ, a $1D$ \texttt{conv} layer with a filter size of 1 is employed in the skip connection to align the input dimension with the output dimension.

The design of the \texttt{ResNet1D} model is depicted in Figure~\ref{fig:resnet_1d}b.
In the figure, each \texttt{block} represents the building block, as shown in Figure~\ref{fig:resnet_1d}a.
The arrow notation, such as $1 \rightarrow 64$ in the first block, indicates the input and output dimensions.
Since our model is specifically designed for univariate time series, the input time series has a dimension of 1.
After passing through the building blocks and the global average pooling layer, the output is a vector of size 128.
The global average pooling operation is applied along the temporal dimension to obtain this output.

To utilize the \texttt{ResNet1D} model for approximating distance functions, we incorporate it as a component within a Siamese neural network architecture~\cite{bromley1993signature}, as depicted in Figure~\ref{fig:resnet_1d}c.
In this setup, we introduce a linear layer after the global average pooling layer to further process the representations generated by the \texttt{ResNet1D} model.
The output dimension of the linear layer is set to 128.
To compute the output distance, we calculate the $l2$ distance between the two 128-sized vectors obtained from the linear layer.

When applying the \texttt{ResNet1D} model to MTL problems, we employ multiple linear layers to compute the logits for each task, as illustrated in Figure~\ref{fig:resnet_1d}d.
Following the conventional hard parameter sharing approach, the parameters of the \texttt{ResNet1D} model are shared among all tasks, while the parameters of the linear layers preceding the \texttt{softmax} layers are task-specific.

\input{insert/fig_resnet_2d}

\subsection{\texttt{ResNet2D} Architecture}
In this section, we first present the proposed \texttt{ResNet2D} model (depicted in Figure~\ref{fig:resnet_2d}a and Figure~\ref{fig:resnet_2d}b) and then discuss its applications in different scenarios.
The design of the \texttt{ResNet2D} model may appear unconventional for processing time series data initially.
However, the rationale behind incorporating $2D$ convolutional layers becomes clearer when we demonstrate how the model is employed in distance function approximation and MTL (refer to Figure~\ref{fig:resnet_2d}c and Figure~\ref{fig:resnet_2d}d).

Figure~\ref{fig:resnet_2d}a presents the building block of the \texttt{ResNet2D}, while Figure~\ref{fig:resnet_2d}b showcases the overall architecture of the \texttt{ResNet2D}.
The design of our \texttt{ResNet2D} model draws inspiration from the original \texttt{ResNet} designed for computer vision tasks~\cite{he2016deep}.
In our model, each input instance is represented as a tensor with dimensions $n \times m \times k$, where $n$ and $m$ refer to the width and height, respectively, while $k$ represents the depth or feature dimension.
The specific values of $n$, $m$, and $k$ may vary depending on the dataset and application at hand.
Our architecture is composed of 8 blocks, and following each block, we incorporate a $3 \times 3$ \texttt{conv} layer with a stride size of 2 to reduce the width and height of the intermediate representation by half.

We employ the architecture depicted in Figure~\ref{fig:resnet_2d}c for distance learning tasks.
In this setup, if the length of the first input time series is denoted as $n$, and the length of the second time series as $m$, the pairwise distance matrix between the two time series can be represented as a tensor with dimensions $n \times m \times 1$.
This matrix captures the Euclidean distance between each data point in one time series and every data point in the other time series.
To construct the distance learning models, we utilize the \texttt{ResNet2D} model shown in Figure~\ref{fig:resnet_2d}b, setting the value of $k$ to $1$.
The pairwise distance matrix plays a vital role as it provides essential information for computing the distance between two time series across all possible warping paths, enabling the \texttt{ResNet2D} model to learn the warping mechanism.
For instance, when $n=m$, the sum of the matrix diagonal corresponds to the Euclidean distance between the two time series.
By mimicking the recurrent function of the \texttt{DTW} algorithm using the \texttt{ResNet2D} model, we can compute the \texttt{DTW} distance between the two time series based on the matrix.

The MTL variant of the proposed \texttt{ResNet2D} model is depicted in Figure~\ref{fig:resnet_2d}d.
In this configuration, we have a single input time series, unlike the distance learning variant.
To compute the pairwise distance matrix, we employ $k$ learnable time series templates of length $m$.
Therefore, if the input time series has a length of $n$, the output of the pairwise distance operation is a tensor with dimensions $n \times m \times k$.
In our design, we use 64 templates (i.e., $k=64$), each with a length of 512 (i.e., $m=512$).
Similar to the design shown in Figure~\ref{fig:resnet_1d}d, the model also includes multiple parallel output layers for each task.
Each output layer has its independent set of parameters specific to its task, while the parameters associated with the core \texttt{ResNet2D} model and the templates are shared across different tasks.

To gain insight into how the \texttt{ResNet2D} model can learn the warping mechanism, let us revisit the algorithms for computing the \texttt{DTW} distance~\cite{rakthanmanon2012searching} and the \texttt{soft-DTW} distance~\cite{cuturi2017soft}.
Both distances can be computed by employing dynamic programming recursion on the pairwise distance matrix between two time series, as depicted in Algorithm~\ref{alg:warping}.
For the \texttt{DTW} distance, the recursion function $\textsc{Recursion}(x_0, x_1, x_2)$ is defined as $\textsc{Min}(x_0, x_1, x_2)$.
On the other hand, for the \texttt{soft-DTW} distance, the recursion function $\textsc{Recursion}(x_0, x_1, x_2)$ is defined as $-\gamma \sum_{i=1}^{3}e^{-x_i/\gamma}$, where $\gamma$ is a hyper-parameter for controlling the behavior of the \texttt{soft-DTW} distance.

\input{insert/alg_warping}

The core \texttt{ResNet2D} model plays a similar role to the $\textsc{Recursion}(\cdot)$ function in both distance algorithms.
With its expressive power, the \texttt{ResNet2D} model is capable of learning various recursion functions, thereby capturing different warping mechanisms.
As demonstrated in Section~\ref{sec:exp:dtw}, the proposed \texttt{ResNet2D} model effectively approximates different variants of the \texttt{DTW} distance functions.

\subsection{Optimization}
The models are optimized using the mini-batch stochastic gradient descent (\texttt{SGD}) algorithm, specifically employing \texttt{AdamW} \cite{loshchilov2017decoupled} as the optimizer.
For the distance function approximation experiments, we utilize the mean squared error between the ground truth and the model's prediction as the loss function. 
In this case, the standard mini-batch \texttt{SGD} can be directly applied without any modifications.
In the time series classification (TSC) experiments, we employ cross-entropy as the loss function. 
However, due to potential variations in the number of time series and the lengths of each time series across different tasks, we have made modifications to the standard mini-batch \texttt{SGD} algorithm to accommodate these differences. 
The modified mini-batch \texttt{SGD} algorithm is outlined in Algorithm~\ref{alg:mtl_sgd}.

\input{insert/alg_mtl_sgd}

The main distinction between Algorithm~\ref{alg:mtl_sgd} and the standard mini-batch \texttt{SGD} algorithm lies in the determination of the number of iterations for each epoch, which is based on the task with the largest dataset (refer to \texttt{lines} 3 to 5).
When the task with a smaller dataset reaches its last example during the construction of mini-batches, the dataset is shuffled, and the mini-batch counter is reset (see \texttt{lines} 13 to 15).
This implies that some or all examples from a task with a smaller dataset may be sampled more than once within a single epoch, whereas examples from the task with the largest dataset are sampled only once per epoch.

It is important to note that all examples within a batch are drawn from the dataset associated with the same task.
This design decision stems from the varying lengths of time series across different tasks.
By enforcing this restriction, we guarantee that all time series within a mini-batch share the same length, facilitating efficient model updates.
Furthermore, to mitigate bias towards any specific task during the updates, the order of tasks is shuffled for each iteration (see \texttt{line} 10).



%% file: insert/fig_resnet_1d.tex
\begin{figure}[ht]
\centerline{
\includegraphics[width=0.99\linewidth]{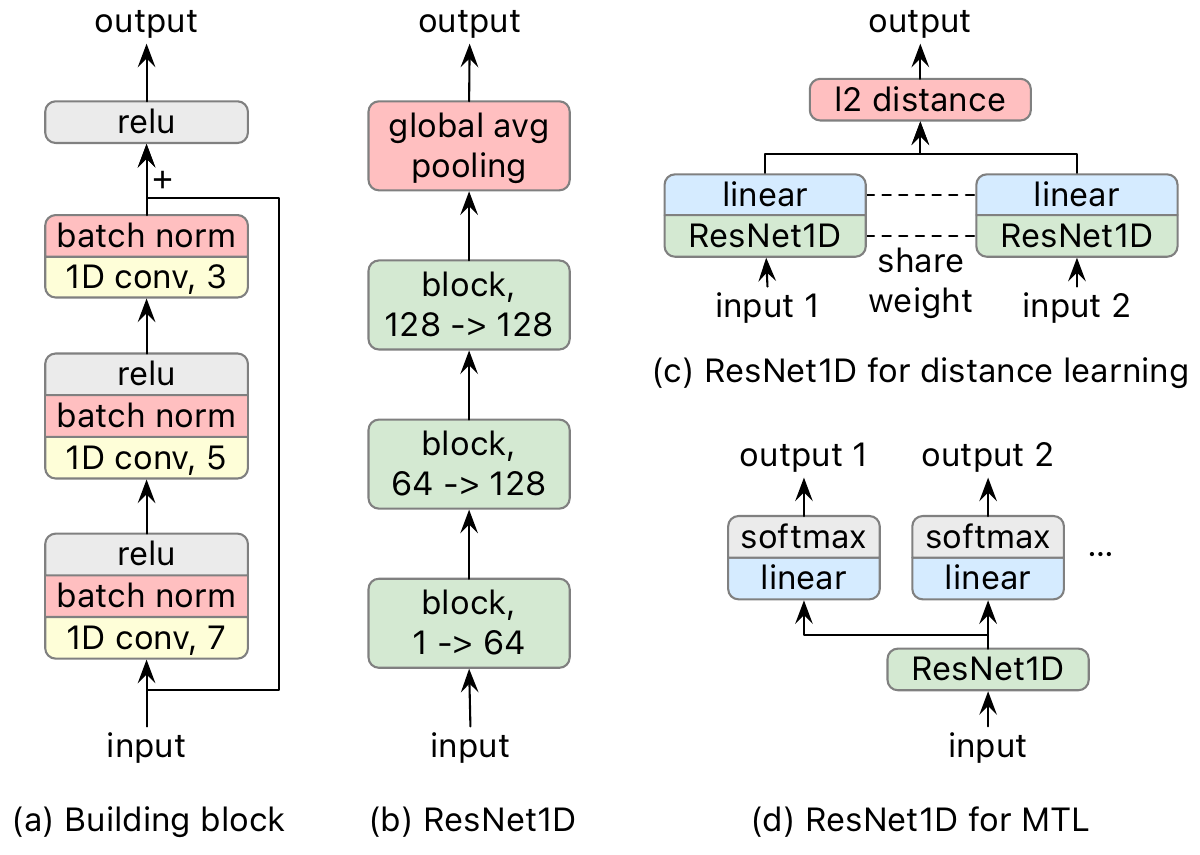}
}
\caption{
The figure illustrates the design of the baseline \texttt{ResNet1D} model and demonstrates how it is utilized for both distance learning and MTL experiments.
}
\label{fig:resnet_1d}
\end{figure}

%% file: insert/fig_resnet_2d.tex
\begin{figure*}[ht]
\centerline{
\includegraphics[width=0.75\linewidth]{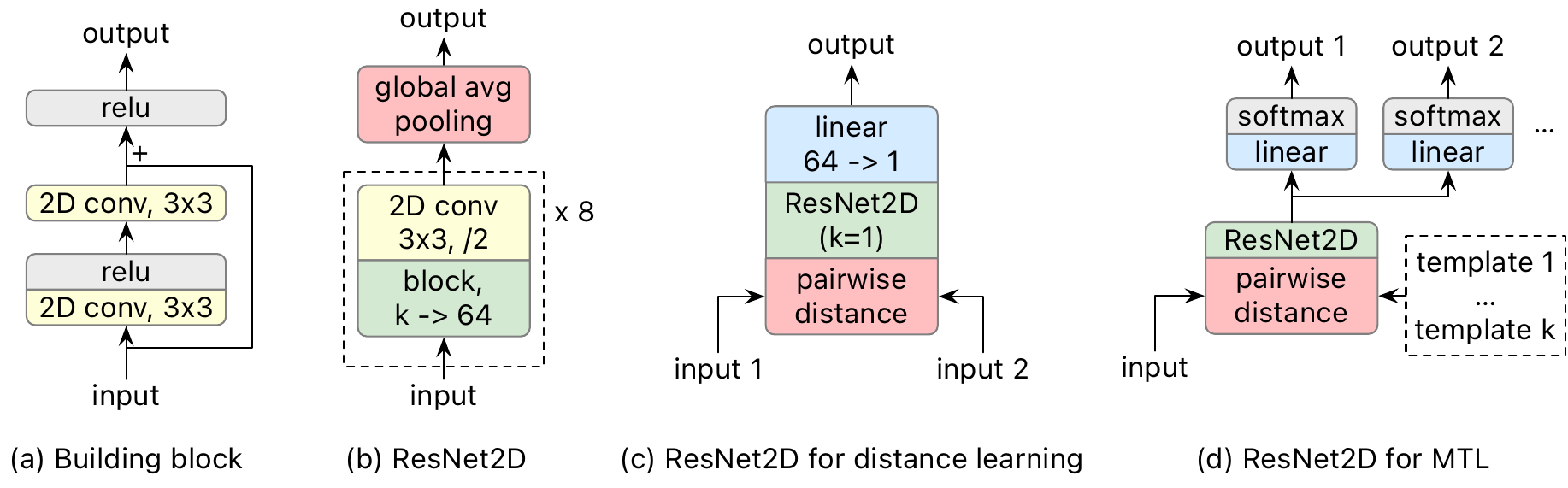}
}
\caption{
The figure showcases the design of the proposed \texttt{ResNet2D} model and illustrates how the model is employed for both distance learning and MTL experiments.
}
\label{fig:resnet_2d}
\end{figure*}

%% file: insert/alg_warping.tex
\begin{algorithm}[ht]
    \centering
    \caption{DTW/soft-DTW Distance\label{alg:warping}}
    \footnotesize
    \begin{algorithmic}[1]
        \Input{pairwise distance matrix~$X \in \mathbb{R}^{n \times m}$}
        \Function{Distance}{$X$}
        \For{$i$ \textbf{in} $[0, \cdots, n)$}
        \For{$j$ \textbf{in} $[0, \cdots, m)$}
        \State $X[i, j] \gets X[i, j] +\textsc{ Recursion}($\par
        \hspace*{5em} $X[i-1, j], X[i, j-1], X[i-1, j-1])$
        \EndFor
        \EndFor
        \State \Return $X[n-1, m-1]$
        \EndFunction
    \end{algorithmic}
\end{algorithm}

%% file: insert/alg_mtl_sgd.tex
\begin{algorithm}[ht]
    \centering
    \caption{Mini-batch SGD for MTL\label{alg:mtl_sgd}}
    \footnotesize
    \begin{algorithmic}[1]
        \Input{datasets~$\mathbf{D}$, batch size~$n_{\text{batch}}$, number of epoch~$n_{\text{epoch}}$, model~$M$}
        \Output{model~$M$}
        \Function{Train}{$\mathbf{D}, n_{\text{batch}}, n_{\text{epoch}}, M$}
        \State $n_{\text{iter}} \gets 0$
        \For{\textbf{each} $D$ \textbf{in} $\mathbf{D}$}
        \State $n_{D} \gets |D| \mathbin{/} n_{\text{batch}}$
        \State $n_{\text{iter}} \gets \textsc{ Max}(n_{\text{iter}}, n_{D})$
        \EndFor
        \For{$i$ \textbf{in} $[0, \cdots, n_{\text{epoch}})$}
        \For{\textbf{each} $D$ \textbf{in} $\mathbf{D}$}
        \State $\textsc{Shuffle}(D)$
        \EndFor
        \For{$j$ \textbf{in} $[0, \cdots, n_{\text{iter}})$}
        \State $\textsc{Shuffle}(\mathbf{D})$ \Comment{shuffle the order of task}
        \For{\textbf{each} $D$ \textbf{in} $\mathbf{D}$}
        \State $X, Y \gets \textsc{ GetNextMiniBatch}(D)$
        \If{reaching the end of $D$} 
        \State $\textsc{Shuffle}(D)$
        \State restart the mini-batch counter for $D$ 
        \EndIf
        \State $M \gets \textsc{ UpdateModel}(X, Y, M)$
        \EndFor
        \EndFor
        \EndFor
        \State \Return $M$
        \EndFunction
    \end{algorithmic}
\end{algorithm}

%% file: section/experiment.tex
\section{Experiment}
We present three sets of experiments in this section.
In the first set of experiments, we explore the difference between the proposed \texttt{ResNet2D} model and the baseline \texttt{ResNet1D} model in terms of model expressiveness, focusing on the \texttt{DTW} distance approximation problem.
Next, we create seven MTL datasets from the publicly available UCR archive~\cite{dau2019ucr} for the second set of experiments.
Using these datasets, we compare the performance of the proposed \texttt{ResNet2D} model with the baselines.
In the last set of experiments, we demonstrate the effectiveness of the proposed \texttt{ResNet2D} model in solving real-world time series classification problems.

\subsection{Distance Function Approximation} 
\label{sec:exp:dtw}
One way to compare the expressiveness of different models is by assessing their ability to approximate a complex function.
In this set of experiments, we train both the baseline \texttt{ResNet1D} model (see Figure~\ref{fig:resnet_1d}c) and the proposed \texttt{ResNet2D} model (see Figure~\ref{fig:resnet_2d}c) to approximate the \texttt{DTW} distance function under various warping constraints.

We conducted the experiments using synthetic time series data.
To train the models, we generated 100,000 pairs of random walk time series as input data and computed the \texttt{DTW} distances under different warping constraints as the target values.
For the validation dataset, we generated 1,000 pairs of random walk time series and their corresponding ground truth distances for model selection.
Additionally, we generated another 1,000 pairs of random walk time series and their associated ground truth distances for testing purposes.
Both models were trained for 200 epochs, and the best model was selected based on the error between the predicted distances and the ground truth distances on the validation data.
The box plots in Figure~\ref{fig:toy} depict the errors of different methods evaluated on the test dataset.
Overall, the proposed \texttt{ResNet2D} model exhibits lower errors compared to the baseline \texttt{ResNet1D} model, indicating its superior ability to approximate the \texttt{DTW} distance function.

\input{insert/fig_path_viz}

To gain insights into the superb performance of the \texttt{ResNet2D} model, we employ the pixel perturbation technique~\cite{wang2019deepvid,wang2021visual} to generate pixel-importance maps for the pairwise distance matrices of various test time series pairs.
The pixel perturbation method evaluates the importance of individual pixels by measuring the change in output before and after disabling specific pixels~\cite{wang2019deepvid,wang2021visual}.
Figure~\ref{fig:path_viz} illustrates the resulting importance maps, where the red lines represent the ground truth warping paths overlaid on the maps.
Notably, the pixels surrounding the ground truth warping paths exhibit greater importance compared to other parts of the matrix.
These pixel-importance maps provide evidence that the proposed \texttt{ResNet2D} model learns the warping mechanism.

\subsection{UCR Archive} 
\label{sec:exp:ucr}
To compare different MTL models, we create seven MTL datasets using the following steps.
First, we select 25 datasets from the UCR archive~\cite{dau2019ucr} that share common sources with others.
Subsequently, we organize the 25 tasks into seven distinct datasets, grouping them based on the source of the associated time series. 
These groupings are importance for MTL, as they enable the meaningful sharing of knowledge among tasks within the same dataset. 
For instance, all Electrocardiography (ECG) tasks are consolidated into a single dataset. 
To ensure sufficient training samples for each task, we partition the dataset associated with each task into three subsets: 60\% for training, 20\% for validation, and 20\% for testing purposes.
The 25 tasks and their groupings are provided in Table~\ref{tab:ucr_task}.

\input{insert/tab_ucr_task}

We have evaluated five different methods using the seven MTL datasets. 
The tested methods are:
\begin{itemize}
    \item \texttt{1NN-DTW}: This method serves as the standard baseline for TSC problems~\cite{dau2019ucr}. 
    It utilizes the \texttt{DTW} distance with a one-nearest-neighbor classifier and treats each task as an independent TSC dataset.
    \item This is the baseline \texttt{ResNet1D} model depicted in Figure~\ref{fig:resnet_1d}. 
    It does not incorporate MTL and treats tasks as independent TSC problems. 
    Evaluating this method allows us to examine the impact of MTL on the baseline \texttt{ResNet1D} model. 
    Since this method does not employ MTL, it employs the standard mini-batch SGD algorithm for model optimization.
    \item \texttt{ResNet2D}: This is the proposed \texttt{ResNet2D} model illustrated in Figure~\ref{fig:resnet_2d}. 
    Similar to the previous method, it does not involve MTL and treats tasks as independent TSC problems. 
    Evaluating this method enables us to analyze the effect of MTL on the proposed \texttt{ResNet2D} model. 
    As with the previous method, it utilizes the standard mini-batch SGD algorithm for model optimization.
    \item \texttt{ResNet1D} w/ MTL: This is the baseline \texttt{ResNet1D} model with MTL, as shown in Figure~\ref{fig:resnet_1d}d. 
    The model is trained using Algorithm~\ref{alg:mtl_sgd}.
    \item \texttt{ResNet2D} w/ MTL: This is the proposed \texttt{ResNet2D} model with MTL, as depicted in Fiugre~\ref{fig:resnet_2d}d.
    The model is also trained with Algorithm~\ref{alg:mtl_sgd}.
\end{itemize}

Each deep learning-based model is trained for 400 epochs, and a checkpoint is saved at the end of each epoch. 
The best checkpoint is selected based on the validation accuracy. 
A batch size of 40 and a learning rate of 0.0001 are utilized. 
The experimental results are displayed in Table~\ref{tab:ucr_task}. 
The average rank is the primary metric for comparison, as it provides insight into the performance of each method across various tasks. 
A smaller average rank indicates superior performance across all tasks in comparison to the other methods.
The best method for each task is highlighted with bold font, and the second best method is underlined.

First, we compare the performance of \texttt{1NN-DTW} and \texttt{ResNet1D}. 
The experimental results align with previous studies~\cite{wang2017time,ismail2019deep}. 
However, when MTL is applied to \texttt{ResNet1D}, we observe a decrease in the average rank. 
Surprisingly, the overall performance of \texttt{ResNet1D} with MTL is even inferior to the standard baseline of \texttt{1NN-DTW}.
When comparing the proposed \texttt{ResNet2D} model with both variants of \texttt{ResNet1D}, we find that it outperforms both of them. 
Moreover, when \texttt{ResNet2D} is combined with MTL, the performance improves further. 
In summary, the methods are ranked from best to worst as follows: \texttt{ResNet2D} with MTL, \texttt{ResNet2D}, \texttt{ResNet1D}, \texttt{1NN-DTW}, and \texttt{ResNet1D} with MTL. 
The proposed \texttt{ResNet2D} model with MTL exhibits superior performance compared to the other methods.



\input{insert/fig_resnet_rep}

We also visualize the learned representations of both MTL models using the $t$\texttt{-SNE} algorithm.
We specifically choose the SemgHand dataset because two of its tasks have a hierarchical relationship, namely gender and subject/identity.
In the task ``SemgHandGenderCh2," the labels are determined based on the gender of the subjects from which the samples are gathered.
In the task ``SemgHandSubjectCh2," the labels are determined based on the identity of the subjects.

The visualization results are presented in Figure~\ref{fig:resnet_rep}.
Subjects 1, 2, and 3 are females, while subjects 4 and 5 are males.
The latent space learned by the proposed \texttt{ResNet2D} model successfully captures the hierarchical relationship between the labels of different tasks.
The samples are effectively separated first based on the gender of the associated subjects and then based on their identities.
In contrast, the baseline \texttt{ResNet1D} model fails to learn this hierarchical relationship.

\subsection{Transaction TSC}
Understanding transaction data is crucial for financial institutions~\cite{yeh2020multi, yeh2020merchant, yeh2021online, yeh2022embedding}. 
Time series derived from transaction data can be utilized to develop classification and regression models for various business applications~\cite{yeh2020multi, yeh2020merchant, yeh2021online}. 
In this section, we conduct experiments on the merchant classification problem as it aids analysts in gaining a better understanding of the behavior exhibited by different merchants. 
We examine merchants from three perspectives: business category (based on the types of services or goods being sold), geographical location (i.e., region), and the extent to which a merchant operates online versus offline (i.e., face-to-face).

We construct the dataset using transactions associated with approximately 87,000 merchants spanning from June 1, 2020, to November 1, 2021. 
For each merchant, we generate a time series by calculating the average number of transactions per hour over a week. 
This results in a time series representation for each merchant with a length of 168 (i.e., 7 days multiplied by 24 hours). 
The merchants are divided into training, validation, and test sets using a 6:2:2 split ratio.

\input{insert/tab_visa_acc}


We conducted experiments on the dataset using three methods: \texttt{1NN-DTW}, \texttt{ResNet1D} w/ MTL, and \texttt{ResNet2D} w/ MTL. 
The non-MTL variants were not evaluated since the focus of this project was on developing an MTL model. 
Each model was trained for 200 epochs, and a checkpoint was saved at the end of each epoch. 
The best checkpoint was selected based on the validation accuracy. 
The models were trained with a batch size of 40 and a learning rate of 0.0001.

The experimental results are presented in Table~\ref{tab:visa_acc}. 
For this dataset, the overall order of the methods from best to worst is as follows: \texttt{ResNet2D} w/ MTL, \texttt{ResNet1D} w/ MTL, and \texttt{1NN-DTW}. 
Taking into consideration the results from both Table~\ref{tab:visa_acc} and Table~\ref{tab:ucr_task}, we can conclude that the proposed method (\texttt{ResNet2D} with MTL) consistently outperforms the other methods across different tasks. 

%% file: insert/fig_path_viz.tex
\begin{figure*}[t]
\centerline{
\includegraphics[width=0.9\linewidth]{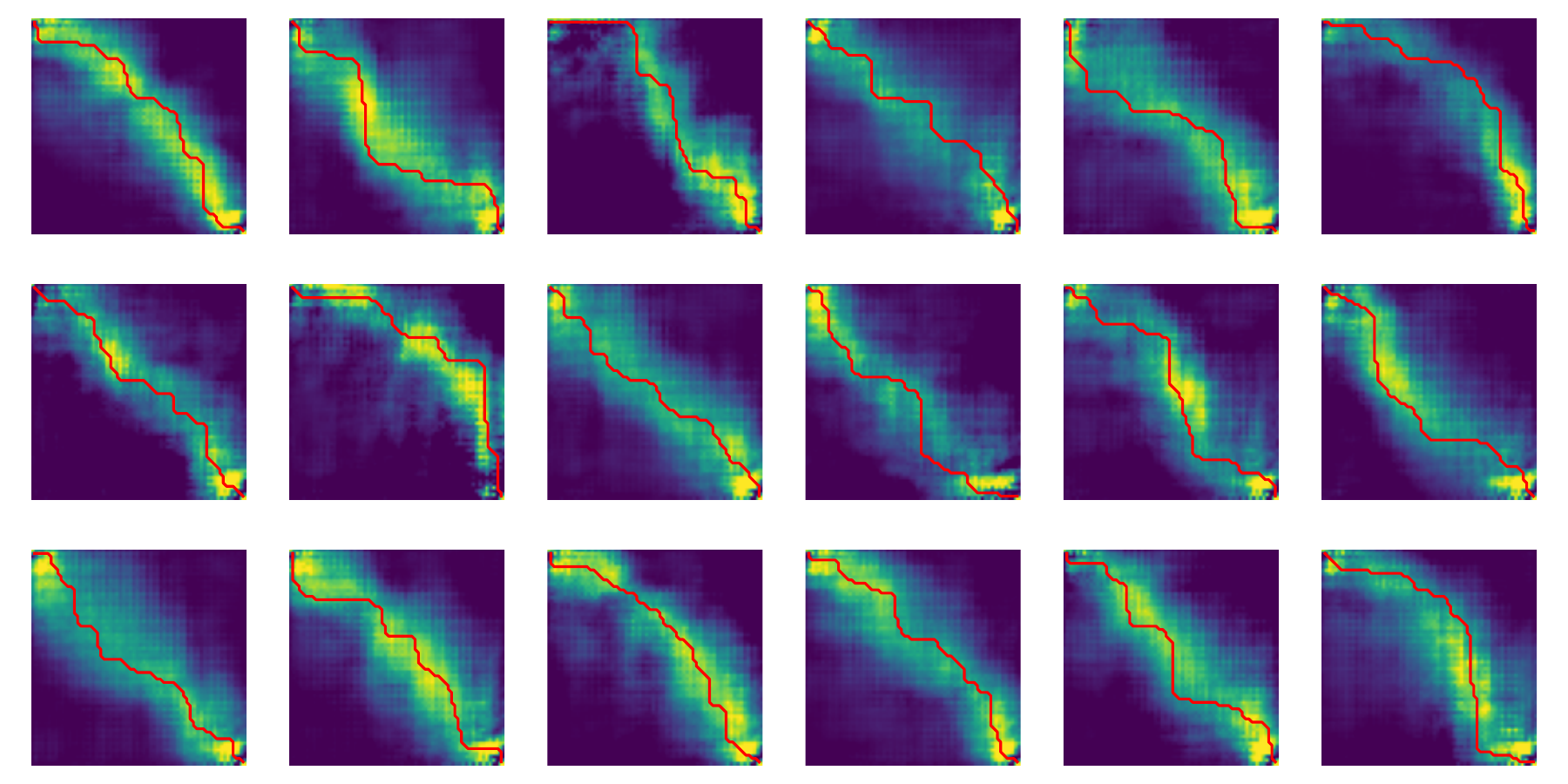}
}
\caption{
The images display the pixel-importance maps for randomly selected pairs of test time series in the proposed \texttt{ResNet2D} model.
On each importance map, we superimpose the ground truth warping path as a red line.
The \texttt{ResNet2D} model can learn the warping mechanism, enabling it to approximate DTW. 
The color gradient from dark blue to orange represents the importance values, with darker shades indicating lower values and brighter shades indicating higher values.
}
\label{fig:path_viz}
\end{figure*}

%% file: insert/tab_ucr_task.tex
\begin{table*}[t]
\centering
\caption{
The 25 selected tasks are organized into seven datasets.
The performance is evaluated using accuracy, and the number within the bracket indicates the rank of each method among the five tested methods.
Among the tested methods, the proposed \texttt{ResNet2D} w/ MTL outperformed the alternatives.
}
\label{tab:ucr_task}
\resizebox{0.85\linewidth}{!}{
\begin{tabular}{ll||c|cc|cc}
Dataset & Task & \texttt{1NN-DTW} & \texttt{ResNet1D} & \texttt{ResNet2D} & \begin{tabular}[c]{@{}l@{}}\texttt{ResNet1D} \\      w/ MTL\end{tabular} & \begin{tabular}[c]{@{}l@{}}\texttt{ResNet2D} \\      w/ MTL\end{tabular} \\ \toprule\toprule
\multirow{3}{*}{Cricket} & CricketX & \underline{0.80 (2.5)} & 0.75 (4.5) & \underline{0.80 (2.5)} & 0.75 (4.5) & $\mathbf{0.95 (1.0)}$ \\
 & CricketY & $\mathbf{0.78 (1.5)}$ & $\mathbf{0.78 (1.5)}$ & 0.71 (4.0) & 0.68 (5.0) & \underline{0.76 (3.0)} \\
 & CricketZ & \underline{0.78 (2.5)} & \underline{0.78 (2.5)} & 0.77 (4.0) & 0.75 (5.0) & $\mathbf{0.94 (1.0)}$ \\ \midrule
\multirow{4}{*}{UWaveGestureLibrary} & UWaveGestureLibraryAll & 0.91 (3.0) & 0.90 (4.0) & $\mathbf{0.98 (1.0)}$ & 0.82 (5.0) & \underline{0.97 (2.0)} \\
 & UWaveGestureLibraryX & 0.77 (5.0) & 0.83 (3.0) & $\mathbf{0.86 (1.0)}$ & 0.79 (4.0) & \underline{0.85 (2.0)} \\
 & UWaveGestureLibraryY & 0.64 (5.0) & 0.73 (3.0) & \underline{0.76 (2.0)} & 0.69 (4.0) & $\mathbf{0.80 (1.0)}$ \\
 & UWaveGestureLibraryZ & 0.70 (5.0) & $\mathbf{0.78 (1.5)}$ & \underline{0.77 (3.0)} & 0.75 (4.0) & $\mathbf{0.78 (1.5)}$ \\ \midrule
\multirow{3}{*}{AllGestureWiimote} & AllGestureWiimoteX & $\mathbf{0.81 (1.0)}$ & \underline{0.79 (2.0)} & 0.76 (3.0) & 0.69 (5.0) & 0.74 (4.0) \\
 & AllGestureWiimoteY & \underline{0.89 (2.0)} & $\mathbf{0.90 (1.0)}$ & 0.77 (4.0) & 0.74 (5.0) & 0.81 (3.0) \\
 & AllGestureWiimoteZ & $\mathbf{0.74 (1.0)}$ & \underline{0.70 (2.0)} & 0.64 (5.0) & 0.66 (4.0) & 0.68 (3.0) \\ \midrule
\multirow{3}{*}{DodgerLoop} & DodgerLoopDay & $\mathbf{0.53 (1.5)}$ & 0.41 (4.5) & $\mathbf{0.53 (1.5)}$ & 0.41 (4.5) & \underline{0.50 (3.0)} \\
 & DodgerLoopGame & $\mathbf{0.97 (2.0)}$ & 0.88 (5.0) & $\mathbf{0.97 (2.0)}$ & $\mathbf{0.97 (2.0)}$ & \underline{0.94 (4.0)} \\
 & DodgerLoopWeekend & $\mathbf{1.00 (2.5)}$ & $\mathbf{1.00 (2.5)}$ & $\mathbf{1.00 (2.5)}$ & \underline{0.97 (5.0)} & $\mathbf{1.00 (2.5)}$ \\ \midrule
\multirow{3}{*}{SemgHand} & SemgHandGenderCh2 & 0.84 (4.5) & 0.84 (4.5) & 0.86 (3.0) & \underline{0.92 (2.0)} & $\mathbf{0.99 (1.0)}$ \\
 & SemgHandMovementCh2 & 0.62 (3.0) & 0.49 (5.0) & \underline{0.65 (2.0)} & 0.58 (4.0) & $\mathbf{0.83 (1.0)}$ \\
 & SemgHandSubjectCh2 & 0.77 (4.0) & 0.74 (5.0) & \underline{0.89 (2.0)} & 0.83 (3.0) & $\mathbf{0.98 (1.0)}$ \\ \midrule
\multirow{3}{*}{GestureMidAir} & GestureMidAirD1 & \underline{0.59 (2.0)} & 0.54 (4.0) & $\mathbf{0.62 (1.0)}$ & 0.50 (5.0) & 0.57 (3.0) \\
 & GestureMidAirD2 & \underline{0.60 (2.5)} & 0.47 (5.0) & 0.56 (4.0) & \underline{0.60 (2.5)} & $\mathbf{0.63 (1.0)}$ \\
 & GestureMidAirD3 & \underline{0.34 (2.0)} & 0.22 (5.0) & 0.24 (4.0) & 0.31 (3.0) & $\mathbf{0.35 (1.0)}$ \\ \midrule
\multirow{6}{*}{ECG} & ECG200 & 0.82 (5.0) & $\mathbf{0.93 (1.5)}$ & \underline{0.90 (3.0)} & 0.88 (4.0) & $\mathbf{0.93 (1.5)}$ \\
 & ECG5000 & \underline{0.94 (4.5)} & $\mathbf{0.95 (2.0)}$ & $\mathbf{0.95 (2.0)}$ & \underline{0.94 (4.5)} & $\mathbf{0.95 (2.0)}$ \\
 & ECGFiveDays & \underline{0.99 (5.0)} & $\mathbf{1.00 (2.5)}$ & $\mathbf{1.00 (2.5)}$ & $\mathbf{1.00 (2.5)}$ & $\mathbf{1.00 (2.5)}$ \\
 & NonInvasiveFetalECGThorax1 & 0.80 (5.0) & $\mathbf{0.94 (1.0)}$ & 0.89 (3.0) & \underline{0.90 (2.0)} & 0.88 (4.0) \\
 & NonInvasiveFetalECGThorax2 & 0.87 (5.0) & $\mathbf{0.94 (1.0)}$ & \underline{0.93 (2.5)} & 0.92 (4.0) & \underline{0.93 (2.5)} \\
 & TwoLeadECG & $\mathbf{1.00 (3.0)}$ & $\mathbf{1.00 (3.0)}$ & $\mathbf{1.00 (3.0)}$ & $\mathbf{1.00 (3.0)}$ & $\mathbf{1.00 (3.0)}$ \\ \midrule
\multicolumn{2}{l||}{Average Rank} & 3.20 & 3.06 & \underline{2.70} & 3.86 & $\mathbf{2.18}$
\end{tabular}
}
\end{table*}

%% file: insert/fig_resnet_rep.tex
\begin{figure}[ht]
\centerline{
\includegraphics[width=0.9\linewidth]{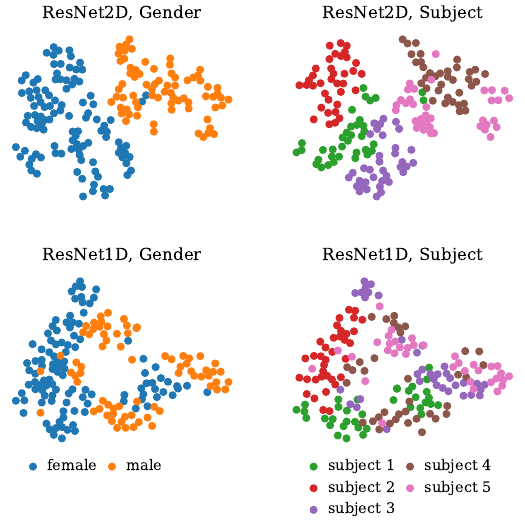}
}
\caption{
The features learned by different MTL models for the SemgHandGenderCh2 and SemgHandSubjectCh2 tasks are visualized using the $t$\texttt{-SNE} algorithm.
Compared to the baseline \texttt{ResNet1D} model, the proposed \texttt{ResNet2D} model demonstrates improved separation of samples from different classes.
}
\label{fig:resnet_rep}
\end{figure}

%% file: insert/tab_visa_acc.tex
\begin{table}[ht]
\centering
\caption{
The proposed \texttt{ResNet2D} w/ MTL outperforms the baseline methods in all three tasks.
}
\label{tab:visa_acc}
\resizebox{0.9\columnwidth}{!}{
\begin{tabular}{l||ccc}
 & Category & Region & Online/Offline \\ \hline \hline
\texttt{1NN-DTW} & 0.2708 & 0.4837 & 0.6629 \\
\texttt{ResNet1D} w/ MTL & 0.2767 & 0.6977 & 0.8388 \\
\texttt{ResNet2D} w/ MTL & \textbf{0.3843} & \textbf{0.7603} & \textbf{0.8407}
\end{tabular}
}
\end{table}

%% file: section/conclusion.tex
\section{Conclusion}
In this paper, we propose the \texttt{ResNet2D} model for time series classification (TSC) under the multitask learning (MTL) setting.
According to the experiments conducted using both publicly available UCR archive \cite{dau2019ucr} and an industrial transaction TSC dataset, the proposed \texttt{ResNet2D} model outperforms its alternatives.
Based on the pixel perturbation visualization technique~\cite{wang2019deepvid,wang2021visual}, the proposed \texttt{ResNet2D} model demonstrates the capability to learn the warping mechanism from time series data. 
This ability is likely a key factor contributing to its superior performance.
For future work, it would be interesting to explore the connection between MTL and the learning process of time series foundation model~\cite{yeh2023toward}.